\documentclass[times,twocolumn,final]{elsarticle}
\usepackage{cag}
\usepackage{framed,multirow}

\usepackage{amssymb}
\usepackage{latexsym}
\usepackage[table,xcdraw]{xcolor}

\usepackage{url}
\usepackage{xcolor}
\definecolor{newcolor}{rgb}{.8,.349,.1}
\usepackage{amssymb}
\usepackage{latexsym}
\usepackage{bbm}
\usepackage{amsmath}
\usepackage{subfigure}
\usepackage{hyperref}

\usepackage{mathtools}
\usepackage{tikz}
\def\checkmark{\tikz\fill[scale=0.4](0,.35) -- (.25,0) -- (1,.7) -- (.25,.15) -- cycle;}

\usepackage[switch,pagewise]{lineno} %Required by command \linenumbers below

\journal{Computers \& Graphics}

\begin{document}

\verso{Preprint Submitted for review}

\begin{frontmatter}

\title{ODFNet: Using orientation distribution functions  to characterize 3D point clouds}%

\author[1]{Yusuf H. Sahin\corref{cor1}} 
\cortext[cor1]{Corresponding author: Yusuf H. Sahin
  }
\ead{sahinyu@itu.edu.tr}
\author[1]{Alican Mertan}
\ead{mertana@itu.edu.tr}
\author[1]{Gozde Unal}
\ead{gozde.unal@itu.edu.tr}

\address[1]{Istanbul Technical University, Computer Engineering, Istanbul, 34469, Turkey }

%\author[1]{First Author Given Name %\snm{Surname}\corref{cor1}}
%\cortext[cor1]{Corresponding author: 
%  Tel.: +0-000-000-0000;  
%  fax: +0-000-000-0000;}
%\emailauthor{example@email.com}{Corresponding Author Name}
%\ead{example@email.com}
    
%\author[2]{Second Author Given Name %\snm{Surname}\fnref{fn1}}
%\fntext[fn1]{Footnote 1.}  

%\address[1]{Address, City, Postcode, Country}
%\address[2]{Address, City, Postcode, Country}

%\received{1 February 2017}
%\received{\today}
%%%% Do not use the below for submitted manuscripts
%\finalform{28 March 2017}
%\accepted{2 April 2017}
%\availableonline{15 May 2017}
%\communicated{S. Sarkar}

\begin{abstract}
Learning new representations of 3D point clouds is an active research area in 3D vision, as the order-invariant point cloud structure still presents challenges for the design of neural network architectures. Recent work explored learning global, local, or multi-scale features for point clouds. However, none of the earlier methods focused on capturing contextual shape information by analyzing local orientation distributions of points. In this paper, we use point orientation distributions around a point in order to obtain an expressive local neighborhood representation for point clouds. We achieve this by dividing the spherical neighborhood of a given point into predefined cone volumes, and statistics inside each volume are used as point features. In this way, a local patch can be represented not only by the selected point's nearest neighbors, but also by considering a point density distribution defined along multiple orientations around the point. We are then able to construct an orientation distribution function (ODF) neural network that makes use of an ODFBlock which relies on MLP (multi-layer perceptron) layers. The new ODFNet model achieves state-of-the-art accuracy for object classification on ModelNet40 and ScanObjectNN datasets, and segmentation on ShapeNet and S3DIS datasets.
\end{abstract}

\end{frontmatter}

%\linenumbers

%% main text
\section{Introduction}

Convolutional neural networks (CNNs) are widely used in vision and pattern recognition problems like object classification, object recognition, and segmentation \cite{lecun2015deep}. However, CNNs have not been applicable to point clouds until recent years. The main obstacle to this was the problem of how to interpret a point in a point cloud representation, which has a permutation-invariant structure, a property not possessed by pixels or voxels in 2D or 3D images. On a 2D or 3D image grid, the convolution operation is defined as a weighted sum in a local neighborhood, which is defined by the kernel. However, in a point cloud, a similar neighborhood structure among the points does not exist. PointNet \cite{qi2017pointnet} pioneered the way to utilizing neural network models for the point cloud classification problem by aiming at constructing a global feature transformation on all the points in the point cloud while respecting their order invariance. While PointNet did not use any neighborhood information, it is argued and shown that using local features improves the performance in recent works \cite{qi2017pointnet++, deng2018ppfnet, xu2018spidercnn}.

\begin{figure}[t]
\begin{center}
   \includegraphics[width=\linewidth]{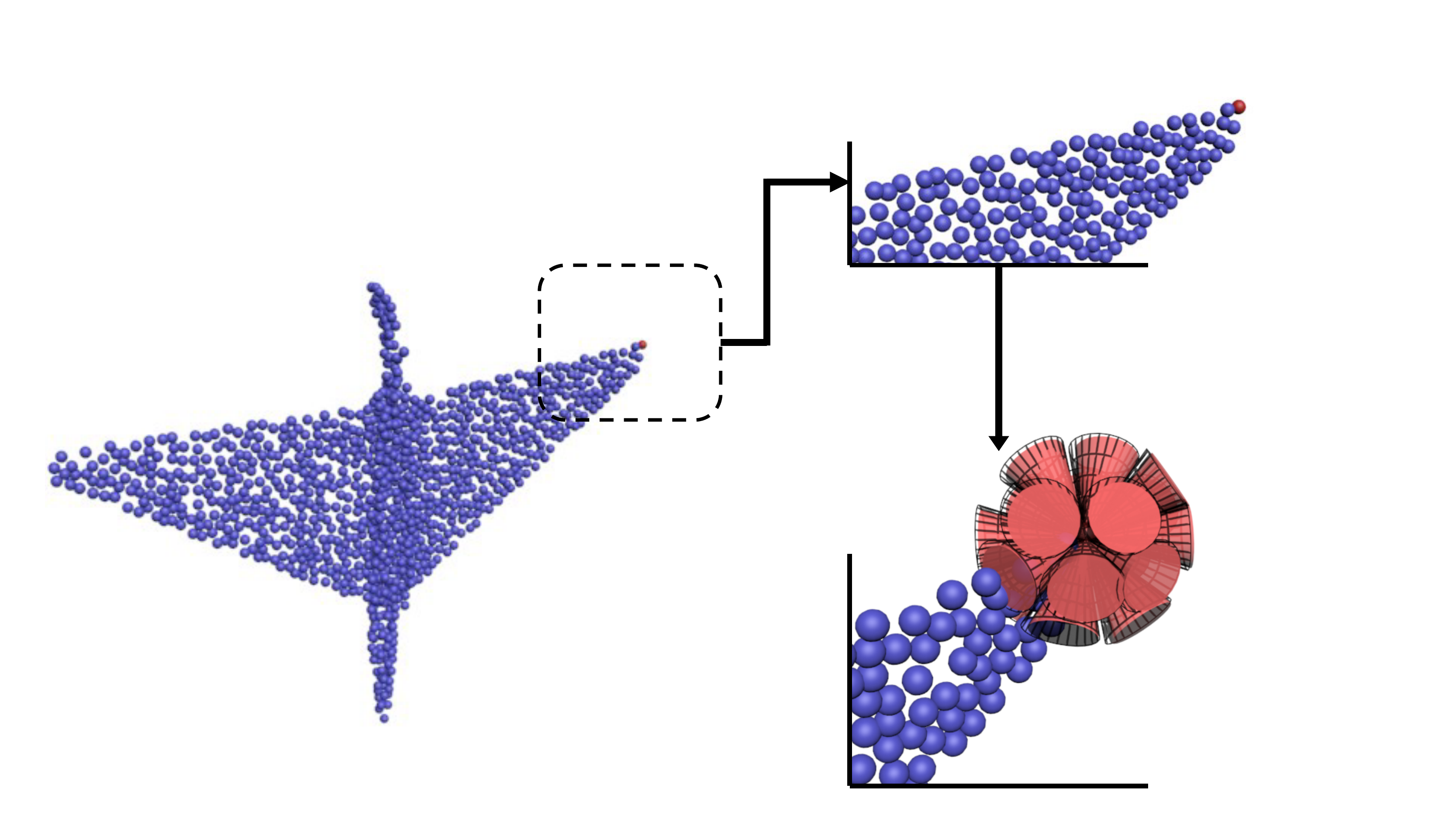}
\end{center}
   \caption{A plane object from the ModelNet40 \cite{wu20153d} dataset. For each point in a point cloud, ODFNet calculates the distribution of nearby points inside a spherical region. To do this, it utilizes cones with predefined orientations that spanning the spherical sectors. These cones can be of different scales and radii.}
\label{fig:Intro}
\end{figure}

Recent approaches to extracting local features from point clouds include creating spherical volumes \cite{qi2017pointnet++} or choosing k-nearest neighbors \cite{wang2019dynamic} and collecting local information of each point in those defined neighborhoods. The latest studies DensePoint \cite{liu2019densepoint} and ShellNet \cite{ zhang2019shellnet}, that obtained state-of-the-art results for classification on the ModelNet40 classification benchmark \cite{wu20153d}, create spherical regions around each point. DensePoint employs spheres of different sizes in each layer of the neural network to obtain features from multiple scales; whereas in ShellNet, coordinates of points in each shell are transformed via an MLP (multi-layer perceptron), and a max-pooling operation aggregates the features across shells. In both methods, the points in the sphere are handled in such a way as to ignore their orientations with respect to the selected point. In \cite{lei2019octree}, a spherical convolution is presented alongside octree partitioning where the sphere is divided into bins, and bin features are obtained by averaging the features of the points inside the bin, without using any point density information.

\begin{figure}[t!]
\begin{center}
\includegraphics[width=\linewidth]{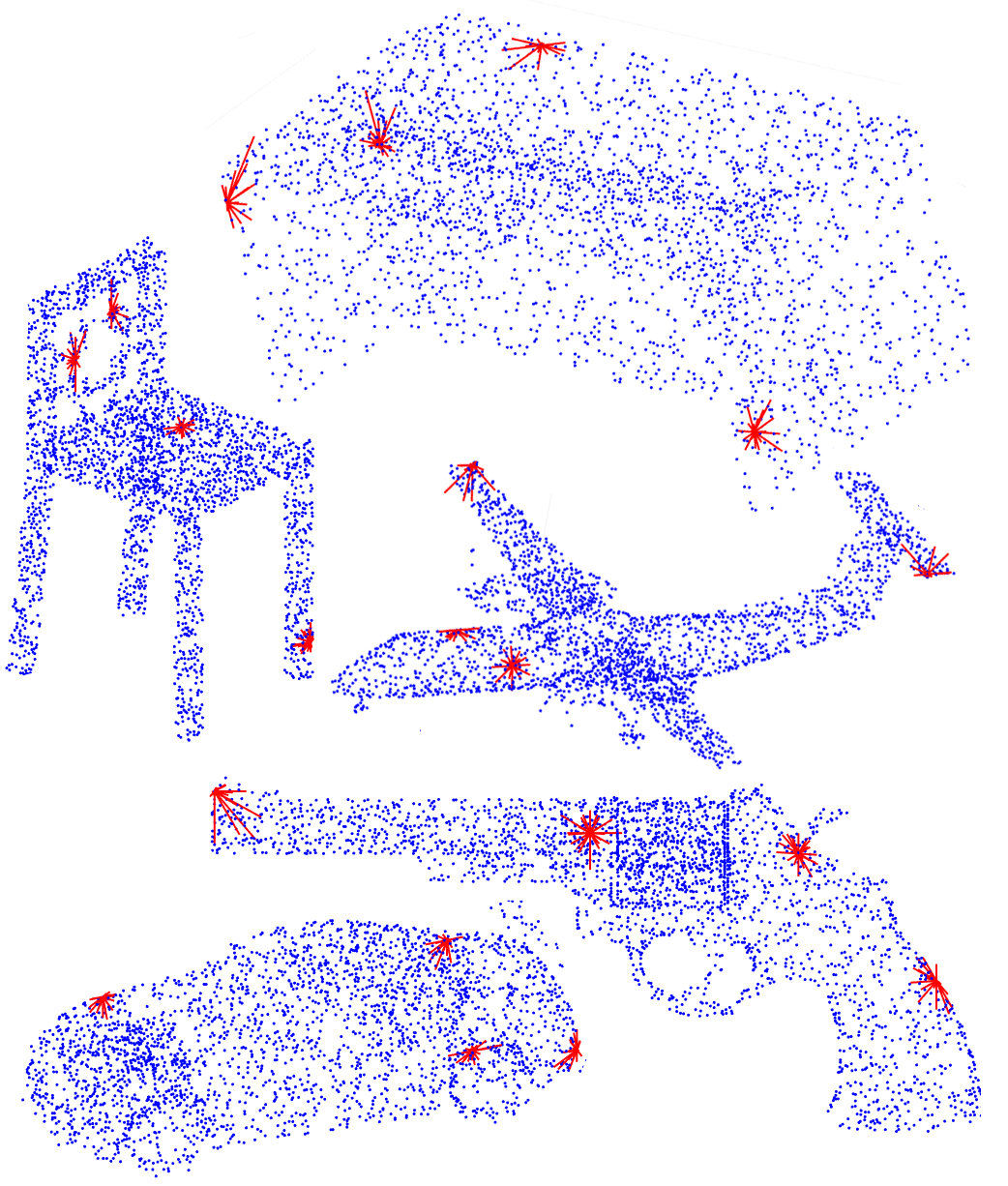}
\end{center}
\caption{ODFs for some selected points on example point clouds from the ShapeNet \cite{yi2016scalable} dataset. Line length indicates strength.}\label{fig:teaser}
\end{figure}

In order to increase the representative power of local features, we employ the distribution of orientations of points in a neighborhood with respect to a reference point. This leads to a new representation named point Orientation Distribution Functions (ODFs) for point clouds. ODFs can be computed by dividing each sphere around a point into a set of cones along predefined orientations, and calculating the density of points in each cone, as depicted in Figure~\ref{fig:Intro}. To increase the representative power of the feature, overlapping cones are also used. Some example ODFs are given in Figure \ref{fig:teaser}. It can be observed that the ODF at the tip of the gun object, the ODFs on the corners of the plane or the table, and on the surface of the car, clearly capture the relative orientation of points with respect to the given center point. As such, we utilize the ODFs with their enhanced capability to compactly summarize the local neighborhood structure of a point cloud to our advantage in our point cloud analysis network model design.

The main contributions of our work can be summarized as such:

\begin{itemize}
    \item The point ODFs, which incorporate the directional information inside a spherical neighborhood, are defined.
    \item A dedicated neural network architecture, the ODFNet, for classification and segmentation of point clouds, is presented.
    \item For different rotation-invariance scenarios, two different ways of utilizing ODFs for a point are presented. Both representations, which are either fully rotation-invariant or only rotation-invariant in the x-y plane, can be used depending on the alignment conditions of the environment.
    \item The ODFNet architecture is tested on popular benchmarks, and state-of-the-art (SoTA) accuracy scores on ModelNet40, Shapenet \cite{yi2016scalable}, S3DIS, \cite{armeni20163d} and ScanObjectNN \cite{uy2019revisiting} are obtained.
\end{itemize}

\section{Related Works}

Earlier studies like \cite{sedaghat2016orientation, li2016fpnn, maturana2015voxnet, wu20153d} focusing on the classification of 3D objects prefer to voxelize the objects and use the voxelized occupancy map as an input to a neural network. However, this approach is not efficient for two reasons: First, the voxelization quality is highly related to the selected grid spacing and, as grid spacing dimensions get lower, distortion of a voxelized 3D object increases. Although high-resolution voxel grids are desirable, they are impractical due to computational constraints. Considering that the input to a voxel grid network is a 3D matrix, the network will consume significantly more storage and computation power when compared to 2D networks. A second drawback is that this is a very sparse representation, hence a superfluous amount of data is unnecessarily processed. \cite{riegler2017octnet} constructed a new representation to decrease the data amount by processing voxels, but the deformation problem remains.

Another approach is obtaining 2D views or depth maps and using them as inputs to a neural network \cite{sfikas2018ensemble, su2015multi,zanuttigh2017deep, guo2016multi}. Feng et al. \cite{feng2019hypergraph} construct hypergraphs extracted from view-based networks. However, view-based approaches are not deemed favorable as well since complete object or scene information is not utilized.

In order to make use of the standard grid convolution operation from the Euclidean CNN domain, Hua et al. project a grid onto every point where filter kernels are placed, and features are calculated over those grids via the convolution operation \cite{hua2018pointwise}. Li et al. \cite{li2018pointcnn} presented an architecture that learns a transformation matrix that weighs and permutes the points to be used in grid convolution.

PointNet \cite{qi2017pointnet} is considered as the first attempt to use point clouds as raw inputs to a neural network. As it might be expected, the main difficulty of using directly the points instead of view renders or voxelized 3D maps comes from the set representation since all permutations on a point set describes the same entry. Hence in PointNet, to classify a point cloud, all points are processed in multi-layer perceptrons in parallel (i.e. as shared weights for all points) to obtain point features and a symmetric function (e.g. a max-pooling operation) is used over these features to obtain an aggregated global feature. Although studies like \cite{hua2018pointwise} showed that ordered points can be used without a symmetric function, the symmetric function notion is widely used \cite{deng2018ppfnet, yang2018foldingnet}. Other studies using PointNet as a backbone network or in the middle steps include  \cite{zamorski2020adversarial, deng2018ppfnet}.

PointNet++ \cite{qi2017pointnet++} focuses on the fact that the original PointNet loses local features since all points are treated independently until the max-pool step. Thus, they hierarchically sample and group the point cloud and implement mini PointNets for each group. In \cite{wang2019dynamic}, DGCNN, which handles the point cloud as a graph and uses a k-nearest neighbor approach to construct the connections, is introduced. Then, an edge convolution operation to perform a convolution centered on the selected point according to its nearest neighbors is defined. In SpiderCNN \cite{xu2018spidercnn}, a new convolution operation that benefits from Taylor series expansion is presented. In KPConv \cite{thomas2019kpconv}, kernel points with learnable weights are defined inside a local neighborhood and a linear correlation between a kernel point position and a neighbor point position is calculated and multiplied by these weights. In DensePoint\cite{liu2019densepoint} and ShellNet  \cite{zhang2019shellnet}, a dedicated convolution is defined relying on  statistics inside local neighborhoods, particularly spherical regions. Lei et al. \cite{lei2019octree} design a procedure where spherical regions are divided into bins and point features for each bin are collected according to bin weights. Then mean of the point features of each bin is used as bin features. In ConvPoint \cite{boulch2020convpoint}, convolution operation differentiates for spatial and feature operations where the spatial operations are done on randomly selected parts. For further reading on point clouds, a detailed survey on this topic can be investigated \cite{guo2019deep}.

The works we examine so far are not invariant to rotation changes. Recently, developing rotation-invariant point features started to attract more attention in the point cloud community. In \cite{rao2019spherical}, the points are mapped on an icosahedral lattice and a new convolution operation to perform on this structure is defined. In \cite{zhang2019rotation}, a rotation-invariant convolution, the RIConv operator is presented to obtain features from distances and angles in a rotation-invariant manner.  Kim et al. presented  RI-GCN which uses graph convolutions hierarchically \cite{kim2020rotation}. In CG-Conv \cite{zhang2020global}, local reference frames are created for each point neighborhood to obtain the local features and the global features are calculated via anchors.

In the point cloud processing literature, the lack of any orientation-specific local feature representation motivated us to propose the ODFNet in this work. The inspiration for ODFs comes from the orientational probability distribution functions in the Diffusion MRI field \cite{tuch2004q, cetin2015higher} that characterize the water diffusion in the brain. Those ODFs model the heterogeneous local tissue micro-structure in order to extract underlying multiple axonal fiber populations. For point clouds, the indirect analogy relies on constructing orientation distributions of the local point cloud mass that can reveal and help resolve the local geometry of the 3D shape along several directions. This is the main motivation in proposing ODFs for characterizing local structure in point clouds. Theoretically, for ODF estimation, one could use mathematical techniques such as spherical harmonics decomposition \cite{bloy2008computing, descoteaux2007regularized}, or for instance fitting von Mises-Fisher distributions \cite{mcgraw2006mises}, in which the latter involves constructing a multivariate normal distribution on the sphere. Alternatively, one could take a purely discrete approach to estimate an ODF through a histogram computation, by binning the local sphere around a given point into conic volumetric sections, as we perform in this work. We present numerical evidence that shows adopting ODFs in point clouds provides mostly competitive and in some cases superior performance among existing point cloud representations in problems of classification and segmentation of point clouds. This supports our conjecture that the inclusion of directional statistics of the local point cloud density leads to an improved localized structural representation and hence provides a performance upgrade.

\section{Method}
\label{method}

\begin{figure}
\begin{center}
   \includegraphics[width=\linewidth]{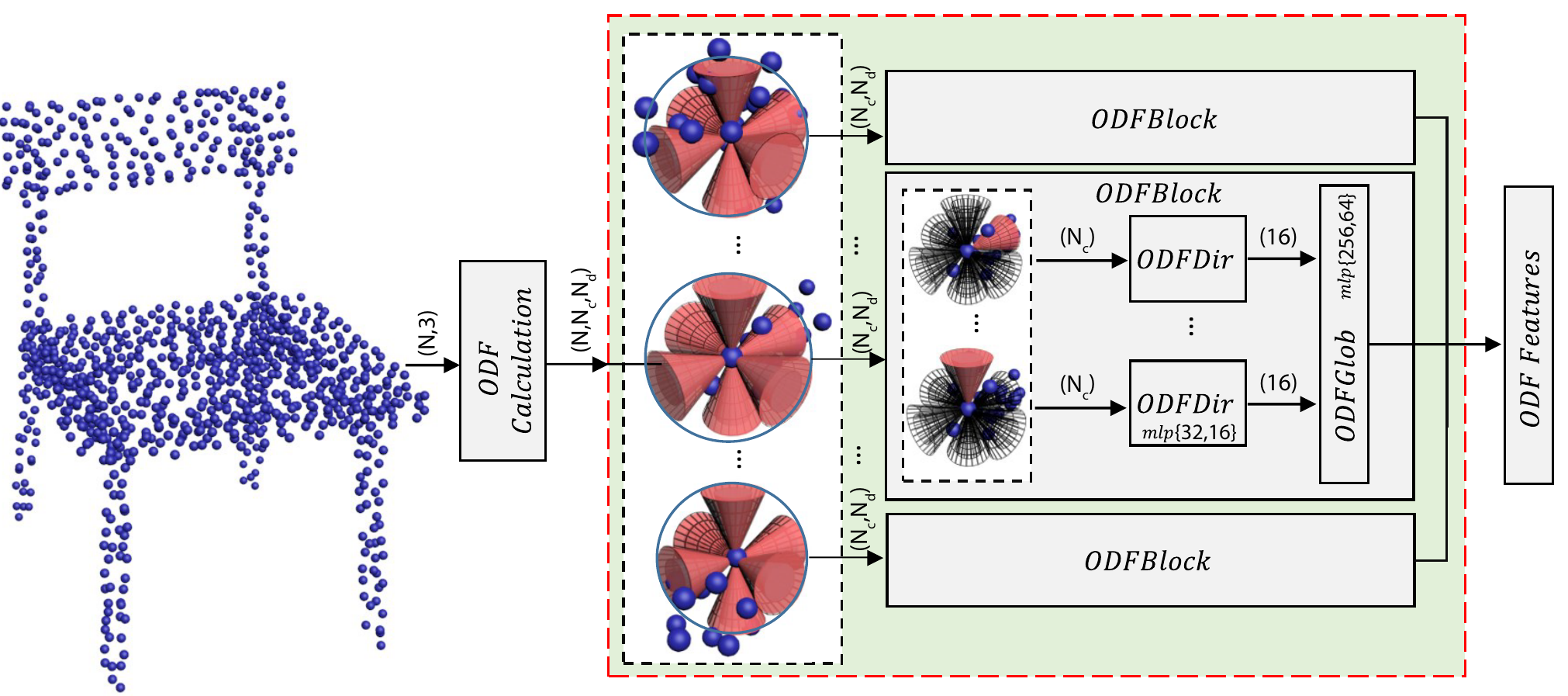}
\end{center}
   \caption{The ODFBlock and its usage to obtain ODF features. Here $N_c$ and $N_d$ represent the number of different cones and directions, respectively. For every point, ODF values are computed along different cones. Then, ODF values are processed in ODFBlocks. An ODFBlock consists of two mlps: ODFDir and ODFGlob.}
\label{fig:ODFBlock}
\end{figure}

Our approach relies upon Orientation Distribution Functions due to their capability to express directional properties of local point cloud structure. In this section, we first describe the ODFs, the dedicated ODFBlock which is depicted in Figure~\ref{fig:ODFBlock}, and then we present the ODFNet.

\subsection{ODF Representation for Point Clouds}

ODFs that we propose in this work rely on the number of points in a local conic neighborhood at multiple orientations. Furthermore, to capture the local point density information in a hierarchy of scales, we utilize multi-scale cones with different apex angles and heights. There are three important components in defining ODFs; namely defining cones, their alignment given a point, and calculation of ODF values.

\begin{figure}[h]
\begin{center}
   \includegraphics[width=0.5\linewidth]{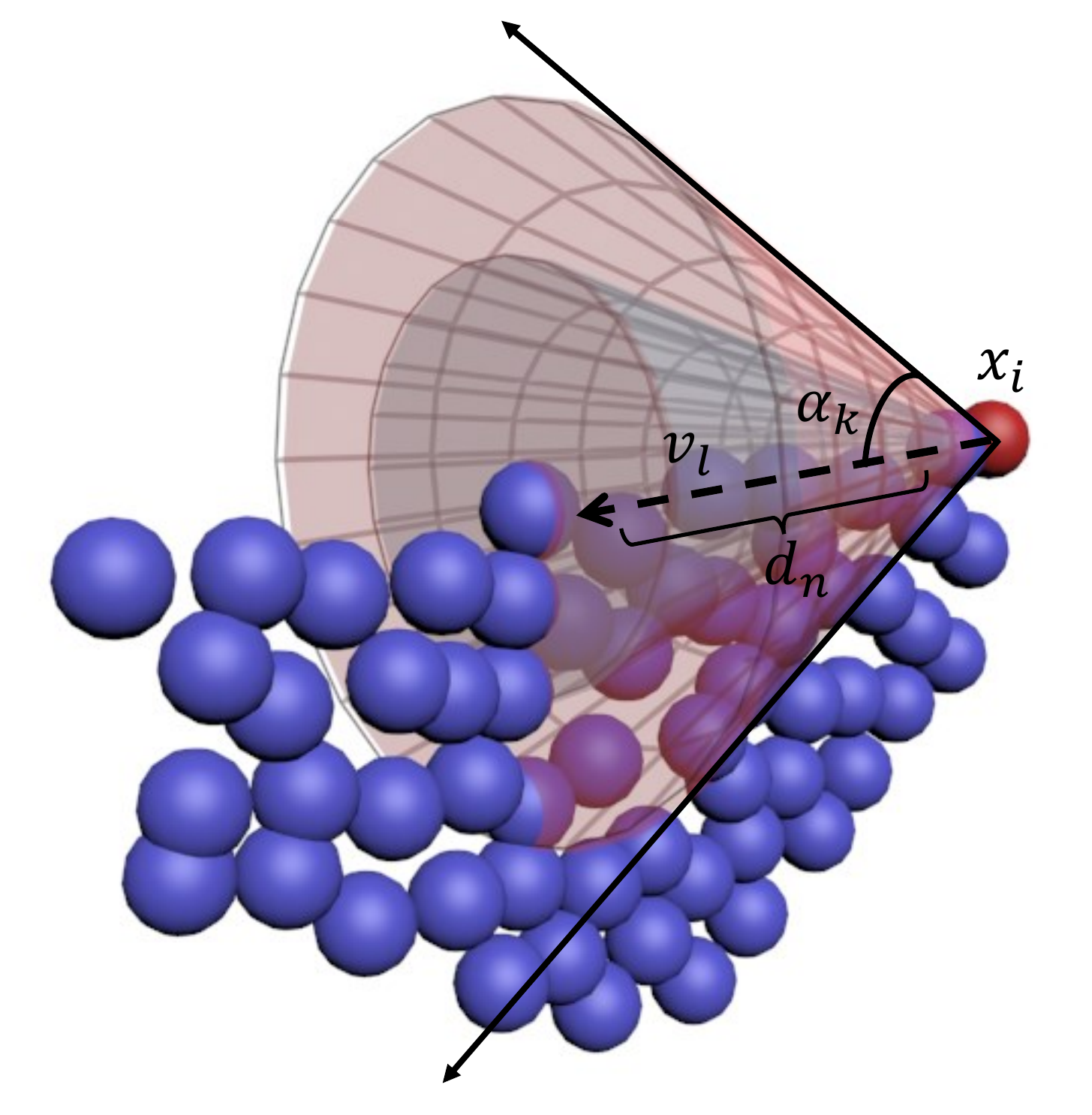}
\end{center}
   \caption{An  example  ODF  cone  is placed  on $x_i$ along direction $v_l$.  Cones at multiple scales (of heights and apex angles) are used to capture point density features at a hierarchy of neighborhoods.}
\label{fig:ODFCone}
\end{figure}

\textbf{Defining cones:} We divide the local sphere around a given point into conical volumes. The motivation for the usage of cones to parcellate the sphere comes from ODF-based methods for the medical imaging field \cite{ehricke2011regularization}. Although there are some studies dividing the sphere using cylinders \cite{cetin2015higher}, a more efficient parcellation in terms of conical volumes is preferred in this work. Each of these cones can be characterized by their direction $v_l$, apex angle $\alpha_k$, and length $d_n$ as illustrated in Figure \ref{fig:ODFCone}. We define $42$ conic neighborhood directions $v_{l}$ for each point,  where directions are obtained after the first tessellation of an icosahedron. This is selected empirically by observing that further increasing the amount of tessellation causes a high rate of intersection between the cones and reducing it decreases the representation power.

For each neighborhood direction $v_l$, we utilize multiple cones to capture features from a hierarchy of neighborhoods. To this end, we use 2 different apex angles $\alpha_{k}$, $31.71$ degrees, which is the smallest angle that covers the whole sphere, and $60$ degrees creating some intersection between the cones; and 4 different distances $d_n$, where $d_n$ is the distance of the $n^{\textsuperscript{th}}$ neighbor of the given point, and $n$ is selected from the collection of $[8,16,24,32]$. Thus, for each point, 336 different cones are obtained (42 cone directions $v_l$, 4 distances $d_n$, and 2 apex angles $\alpha_{k}$).

\begin{figure}[h]
     \centering
     \subfigure[Unaligned]{\includegraphics[width=0.15\textwidth]{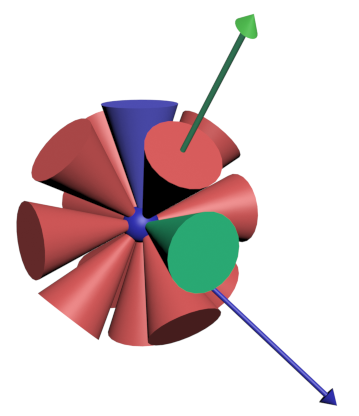}}
     \subfigure[Aligned]{\includegraphics[width=0.15\textwidth]{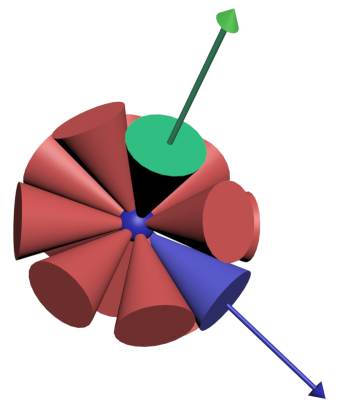}
     }
     \caption{An example alignment of the neighborhoods for a given point. The same rotation that aligns the two neighborhood directions, directions of the blue and the green cone, with the two pivot directions, the blue and green arrows, is applied to all neighborhood directions.}
     \label{fig:aligning_compass}
\end{figure}

\textbf{Aligning neighborhoods for a given point:} In order to make the proposed ODF representation more adaptive to orientation changes of the objects, in the calculation of the ODFs, pivot directions need to be selected. Aligning cone directions $v_l$ according to pivot directions calculated for each point as in Figure \ref{fig:aligning_compass} effectively makes the ODF representation robust to orientation changes of the objects.

Particularly, two orthogonal directions have to be specified to achieve rotation-invariant ODF representations. We devise two methods for selecting these directions,  namely RI-XY and RI-XYZ.

\begin{figure}
\begin{center}
   \includegraphics[width=\linewidth]{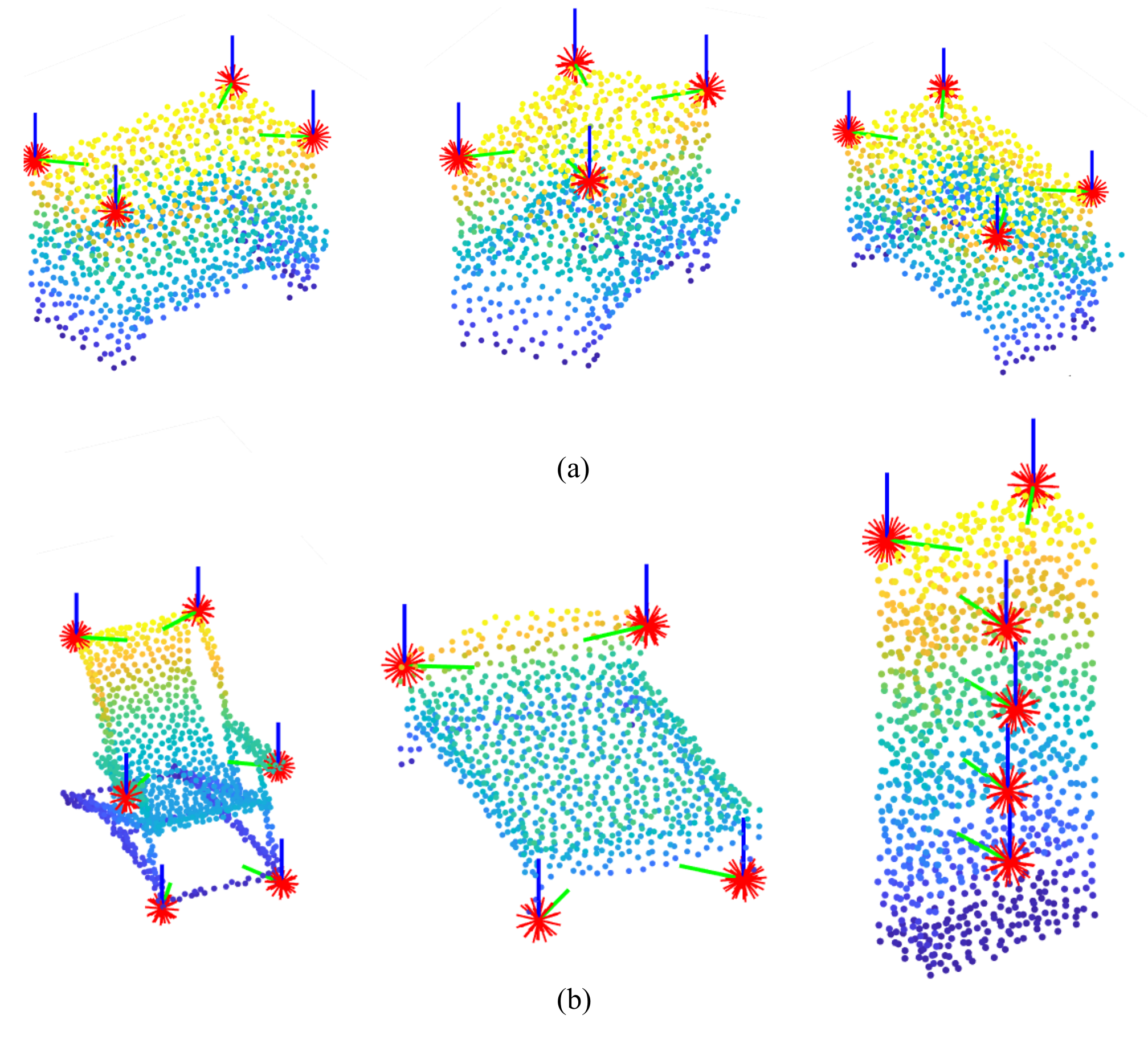}
\end{center}
   \caption{For the RI-XY method, ODF directions according to nearest neighbors are depicted. Green arrows indicate pivot and blue arrows indicate the z-axis. (a) By rotating the objects in the x-y plane, ODF values do not change since the directions are aligned with respect to rotation-invariant pivots. (b) For symmetric points of the same object, nearly symmetric pivot directions are obtained.}
\label{fig:compass}
\end{figure}

RI-XY achieves rotation-invariance in the x-y plane. Thus, it can be used when the object is aligned according to the z-axis. For RI-XY, for every point, projections of the selected point's 32 nearest neighbors on the x-y plane are calculated and the densest direction is selected as the pivot. Using the pivot direction and the z-axis, the ODF directions are aligned. In Figure \ref{fig:compass}, some pivot directions are shown for the RI-XY method. In addition to this alignment's contribution to obtain a rotation-invariant representation for rotations in the x-y plane (Figure \ref{fig:compass}.a), it also leads to symmetric representations for symmetric points (Figure \ref{fig:compass}.b). Unless otherwise stated, the RI-XY method is used for the experiments in this paper.

\begin{figure}
\begin{center}
   \includegraphics[width=0.6\linewidth]{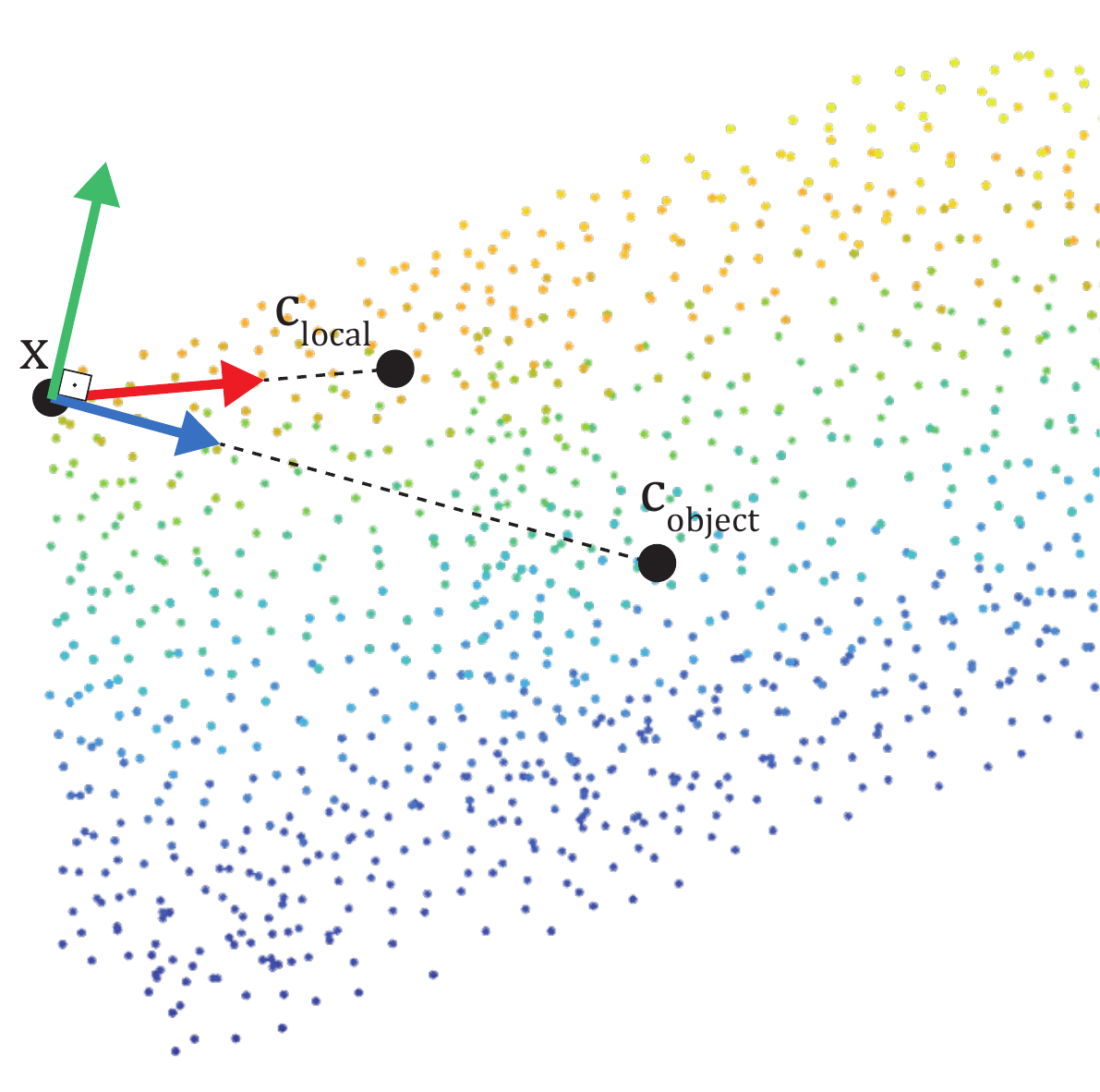}
\end{center}
   \caption{An example of rotation-invariant pivot calculation in RI-XYZ. The blue arrow shows the direction to the object center which is the first pivot, the red arrow shows the direction to the center of the neighbors, and the green arrow is the second pivot direction which is the cross product of the other two vectors.}
\label{fig:pivot_selection}
\end{figure}

RI-XYZ allows us to define fully rotation-invariant representations. For RI-XYZ, the direction from the selected point $x$ to the object center $c_{object}$ is chosen as the first pivot direction. The second pivot direction is the cross product of the first pivot and the vector from point $x$ to the center of the 32 nearest neighbors $c_{local}$ as depicted in Fig \ref{fig:pivot_selection}. This pivot selection technique can be applied when no information about the object's alignment is available. Thus, RI-XYZ could be used for scenarios with totally unaligned objects. However a decrease in performance is expected since the relative coordinates of the points are vastly changed.

 \textbf{Calculation of ODF values:} To compute the ODF value at a point $x_i \in S$, where $S$ denotes the set of all points, and for a specific cone with an apex angle $2\alpha_k$, height $d_n$, and center direction vector $v_l$, 
\begin{multline}\label{eq:ODF}
  ODF(x_i, \alpha_k, d_n, v_l) = \\ 
 \sum_{x_j \in S, i \neq j}\!\mathbbm{1}(||x_i-x_j||_2 < d_n)
\cdot\mathbbm{1}( acos\left(\frac{(x_i-x_j) \!\cdot v_l}{||x_i-x_j||||v_l||}\! \right) \!\!< \!\alpha_k)  
\end{multline}
 which gives us the point count inside the selected cone. Here, $\mathbbm{1}(\cdot)$ refers to the indicator function. Since the cone heights $d_n$ are selected according to the  $n^{\textsuperscript{th}}$-neighbor distance, this value is then normalized by $n$. In our experiments on a single NVIDIA Titan RTX graphics card, calculation of the cone values takes under $\sim76$ msecs for a point cloud of 1024 points despite its complex information. It is also advantageous that the representation is calculated only once in the network.

\subsection{ODFBlock}

In a point cloud, those differently sized and oriented cones that we construct help capture local variations in terms of point density, and encode them into the ODF features, which is provided into the dedicated neural network block in Figure~\ref{fig:ODFBlock}.
After calculating the ODFs at each point, we define the ODFBlock which operates on the ODFs as follows:
\begin{equation}
\!\!\!ODFBlock(x_i, \theta_d, \theta_g) = ODFGlob\left(\theta_d, ODFDir\left(\theta_g, ODF(x_i) \right)\right)
\end{equation}
where $ODF(x_i)$ in short denotes a tensor of point density values $ODF(x_i, \alpha_k, d_n, v_l)$ for different cones along a collection of direction vectors in $\mathbb{S}^2$. Parameters $\theta_d=[\theta^{1}_{d},\theta^{2}_{d},...,\theta^{m}_{d}]$, and $\theta_g=[\theta^{1}_{g},\theta^{2}_{g},...,\theta^{p}_{g}]$ are learnable parameters of the ODFBlock.  $ODFDir$ is an mlp that embeds the ODF tensor through aggregating features by collecting features of each direction, and $ODFGlob$ is another mlp which aggregates the output embedding over all directions to obtain the aggregate ODFBlock output features.

\subsection{ODFNet Network Models}

\subsubsection{ODFNet}

\begin{figure*}[h]
\begin{center}
   \includegraphics[width=\linewidth]{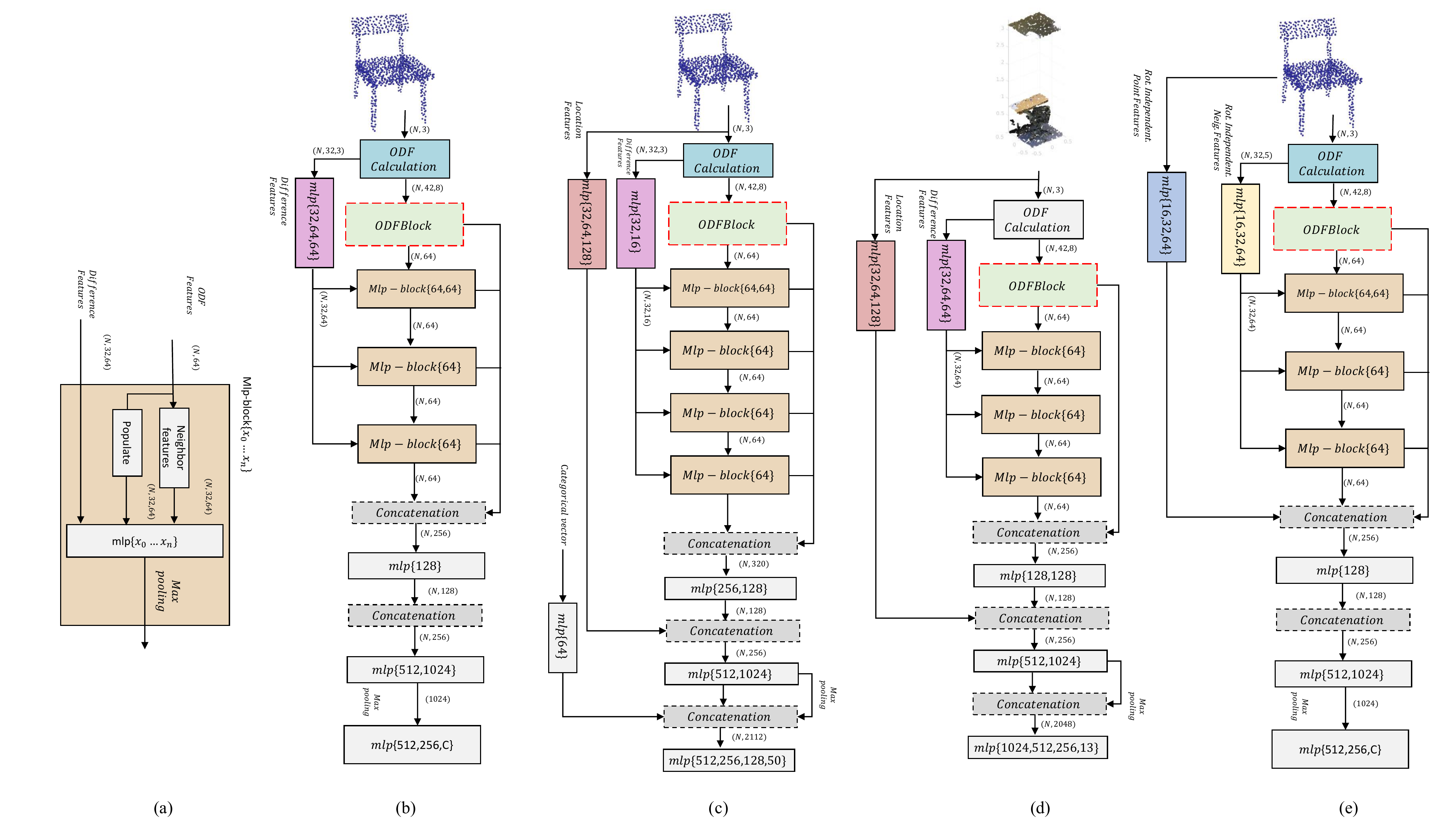}

\end{center}
   \caption{(a) Mlp-block common structure in the ODFNet that is employed for tasks of (b) classification, (c) part segmentation (d) Scene segmentation. (e) represents the ODFNet-XYZ which is fully rotation-invariant. $c$ represents class count.}
\label{fig:ODFarch}
\end{figure*}

To exploit our point ODF's representation capability over point clouds, we design different ODFNets, which benefit from ODFs as well as point locations to capture both local and global features of point clouds and, which can be used for classification and segmentation tasks. As depicted in Figure \ref{fig:ODFarch}, ODFNet first calculates the ODFs for each point as given by Equation~\ref{eq:ODF} for $N_c$ cones rotated around $N_d$ direction vectors. Then for each point, it benefits from two different mlps inside ODFBlock: ODFDir and ODFGlob. For differently scaled cones placed along the same direction, ODFDir calculates features capturing the point density along that direction. ODFGlob then operates on the features captured by ODFDir in order to fuse those features and the result is later concatenated with the point coordinates. 

In the blocks shown as \textit{ODFBlock} in Figure \ref{fig:ODFarch}, in a similar manner with DGCNNs \cite{wang2019dynamic}, each point's features are combined with its nearest neighbors' features and the difference vector features, where the latter represents $\mathbf{x}-\mathbf{x}_i$ for $\mathbf{x}$ indicating a point location and $\mathbf{x_i}$ indicating one of its 32 neighbors.

For the classification task, the last layers of the architecture include a max-pooling operation to produce a global feature vector describing the shape. After three fully connected layers, class scores are obtained at the output. For part segmentation in ShapeNet, the ODFNet architecture makes use of the categorical vector, which indicates the one-hot-coded object class. It is fed to the segmentation part of the ODFNet, as in \cite{wang2019dynamic} and \cite{qi2017pointnet}. The output size of the final output layer in both tasks depends on the number of object classes and object parts. For semantic segmentation in S3DIS, nearly the same architecture with part classification is used. However, since the dataset also includes RGB color information, colors and color differences are also used to obtain difference features and location features.

\subsubsection{ODFNet-xyz}

To ensure the global invariance, we used RI-XYZ pivot selection method and slightly changed the architecture as shown in Figure \ref{fig:ODFarch}.e to obtain ODFNet-xyz, where instead of difference vectors we used a vector set of magnitude of difference vectors $|\textbf{x}-\textbf{x}^i|$, point's distances to object center $|\textbf{x}-\textbf{c}_{object}|$, point neighbors' distances to object center $|\textbf{x}^i-\textbf{c}_{object}|$, angles between points and their neighbors $\angle(\textbf{x},\textbf{x}^i)$, and angles between points and object center $\angle(\textbf{x},\textbf{c}_{object})$. Also $\angle(\textbf{x},\textbf{c}_{object})$ and $|\textbf{x}-\textbf{c}_{object}|$ are concatenated to the features from the last block of the network.

\section{Experimental Results}

In this section, we present the details of the experiments and performance evaluation results for ODFNet on classification and segmentation tasks \footnote{The source code for our ODFNet model will be provided at the time of publication.}. We select widely used benchmark datasets and evaluate the ODFNet model over those, in order to provide a comparison with the existing methods. Particularly, we experiment and report results on ModelNet40 and ScanObjectNN for classification, ShapeNet for part segmentation, and S3DIS for scene segmentation.

\subsection{Shape Classification}

For shape classification, we evaluated our model on ModelNet40 \cite{wu20153d} which consists of mesh models for 40 different categories, and ScanObjectNN \cite{uy2019revisiting} which contains 3D real-life scans for 13 different object categories which also contains noise and missing parts.

\textbf{ModelNet:} To have a fair comparison, we use the preprocessed data from \cite{qi2017pointnet} and use 1024 points for each object. Following the previous studies \cite{wang2019dynamic, qi2017pointnet}, the objects are fit into a unit sphere. Recent works prefer to do scaling and translation \cite{liu2019densepoint}, scaling and perturbing \cite{wang2019dynamic}, and only perturbing \cite{zhang2019shellnet} for data augmentation during training. Instead, nonuniform scaling, flipping in x and y directions, and rotation by multiples of 90 degrees are applied. Also, random sampling is used as in \cite{li2018pointcnn} by deleting half of the points before the last classification block before max-pooling. Since the classification features are obtained by a max-pooling layer, the deletion operation does not affect the network structure. The ODFs are calculated regardless of this operation to avoid shape inconsistencies. To make a fair comparison with the previous work, we compared our study with the methods that use only 1024 points for each object. The results in Table \ref{table:MODELNETScoreTable} show that our implementation outperforms other methods by an overall accuracy score of $93.4\%$ via a  single prediction. Also, by applying a voting mechanism with random scaling and averaging the predictions, we obtained state-of-the-art results among the studies that follow a similar voting procedure.

\begin{table}
\centering
\caption{Overall Accuracies (OA) for point cloud classification results on ModelNet40 dataset. (\textbf{p:} points, \textbf{n:} normals)}
\label{table:MODELNETScoreTable}
\centering
\begin{tabular}{ llll }
\hline
Method  & input & voting & OA                             \\ \hline
PointNet \cite{qi2017pointnet}            & p   &  & 89.2                           \\
PointNet++ \cite{qi2017pointnet++}        & p+n &  & 90.7                           \\
SpiderCNN \cite{xu2018spidercnn}          & p+n &  & 92.4                           \\
Point2Seq \cite{liu2019point2sequence}    & p   &  & 92.6                           \\
InterpCNN \cite{mao2019interpolated}      & p   &  & 93.0                           \\
PointwiseCNN \cite{hua2018pointwise}      & p   &   & 86.1                           \\
ShapeContextNet \cite{xie2018attentional} & p   &  & 90.0                           \\
KCNet \cite{shen2018mining}               & p   &  & 91.0                           \\
PointCNN \cite{li2018pointcnn}            & p   &  & 92.2                           \\
RS-CNN \cite{liu2019relation}             & p   &  & 92.4 \\
ShellNet \cite{zhang2019shellnet}         & p   &  & 93.1                           \\
DGCNN \cite{wang2019dynamic}              & p   &   & 92.9                           \\
ODFNet                                    & p &  & \textbf{93.4} \\ \hline
Kd-network \cite{klokov2017escape}        & p   & \checkmark & 91.8                           \\
GDANet \cite{xu2020learning}              & p  & \checkmark  & 93.8                           \\
DensePoint \cite{liu2019densepoint}       & p   & \checkmark & 93.2                           \\
RS-CNN \cite{liu2019relation}             & p   & \checkmark & 93.6 \\
ODFNet                                    & p & \checkmark & \textbf{94.2} \\ \hline

\end{tabular}
\end{table}

Using the RI-XYZ pivot to calculate the ODFs, and ODFNet-xyz network architecture, we also evaluated our ODF features in three different scenarios focusing on rotation invariance:  train and test with z rotations (z/z), train and test with SO3 rotations (SO3/SO3), train with z rotations and test with SO3 rotations (z/SO3).

In Table \ref{table:rotations}, we examined the results for these experiments in three groups of works (separated by horizontal lines in the table). The first group of networks \cite{maturana2015voxnet, qi2016volumetric, esteves2018learning, su2015multi, qi2017pointnet, qi2017pointnet++, li2018pointcnn, liu2019relation} which indicates the rotation-variant networks, for the z/z scenario, generally achieves scores that are comparable to their scores for the default setup. However, they lack robustness and their performance drops drastically for SO3/SO3 and z/SO3 scenarios. For the second group of approximately rotation-invariant networks, small standard deviations of accuracy are obtained for different scenarios. The third group, which also includes the ODFNet-xyz, is fully rotation-invariant as verified by the zero standard deviation values. The performances of rotation-invariant methods are consistent across the three scenarios. According to the scores, the ODFNet achieves the second-best results for the more challenging SO3/SO3 and z/SO3 scenarios among totally rotation-invariant methods.

Despite ODFNet's success for the z/z scenario and the original scenario (Table \ref{table:MODELNETScoreTable}), its accuracy is drastically decreased for the z/SO3 scenario. It is a natural result of depending on the object's alignment to the XY-plane for both the training procedure and the architecture.

\begin{table}[]
\caption{ Comparisons of the classification accuracy under different rotation settings. Best results are bolded and second bests are underlined. \textit{std.} represents the standard deviation of accuracy between different settings. The ODFNet-xyz architecture is only used for these experiments.}
\label{table:rotations}
\centering
\begin{tabular}{c|ccc|c}
Method     & z/z          & SO3/SO3 & z/SO3 & std. \\ \hline
VoxNet \cite{maturana2015voxnet}           & 83.0         &87.3   & -     & 3.0 \\
SubVolSup \cite{qi2016volumetric}      & 88.5         & 82.7    & 36.6  & 28.4 \\
SphericalCNN \cite{esteves2018learning}   & 88.9         & 86.9    &78.6  &  5.5 \\
MVCNN 80x \cite{su2015multi}      & 90.2         & 86.0    &81.5  & 4.3 \\
PointNet \cite{qi2017pointnet}        & 87.0         & 80.3    & 21.6  & 41.0 \\
PointNet++ \cite{qi2017pointnet++}       & 89.3         & 85.0    & 28.6  & 33.8 \\
PointCNN \cite{li2018pointcnn}         & \underline{91.3} & 84.5    & 41.2  & 27.2 \\
RS-CNN \cite{liu2019relation}          & 90.3         & 82.6    & 48.7  &  22.1 \\
ODFNet  &     \underline{91.3}         &  88.0       &    17.4 & 34.0  \\
\hline
RIConv \cite{zhang2019rotation}         & 86.5         & 86.4    & 86.4  & \underline{0.1} \\
SPHNet \cite{poulenard2019effective}           & 87.0         & 87.6    & 86.6  &  0.5 \\
SFCNN \cite{rao2019spherical}           & \textbf{92.3}         & \textbf{91.0}    & 85.3  &  3.5 \\ 
\hline
ClusterNet \cite{chen2019clusternet}      & 87.1         & 87.1    & 87.1  & \textbf{0.0} \\
RI-GCN  \cite{kim2020rotation}        & \textit{91.0}         & \textbf{91.0}    & \textbf{91.0}  & \textbf{0.0} \\
GCANet  \cite{zhang2020global}         & 89.0         & 89.2    & 89.1  & \textbf{0.0} \\ 
ODFNet-xyz  &     90.2         &  \underline{90.2}       &    \underline{90.2} & \textbf{0.0}  \\
\hline
\end{tabular}
\end{table}%TODO: cog

\textbf{ScanObjectNN: } We use the original dataset that has 2048 points for each object to have a fair comparison.
To evaluate the ODFNet on ScanObjectNN, an augmentation procedure similar to the ModelNet experiments is used. However, because the object point counts are larger here, 1024 points are randomly selected at train time.
There are five different tasks: OBJ\_ONLY, OBJ\_BG,  PB\_T25, PB\_T25R, PB\_T50R, and PB\_T50RS. In OBJ\_BG, objects with background noise are classified. OBJ\_ONLY consists of objects having no background noise. The other sets are augmented versions of OBJ\_BG. Comparing our results with the scores given in \cite{uy2019revisiting}, the ODFNet model produces SoTA accuracy scores as can be observed in Table \ref{table:ScanaObjectNNScoreTable}. 

\begin{table*}[]
\centering
\caption{Classification Accuracies for different tasks in ScanObjectNN \cite{uy2019revisiting} dataset.}
\label{table:ScanaObjectNNScoreTable}
\centering
\begin{tabular}{l|llllll}
Model  & OBJ\_BG &PB\_T25 & PB\_T25R & PB\_T50R & PB\_T50RS & OBJ\_ONLY \\ \hline
3DmFV \cite{ben20183dmfv}      & 68.2          & 67.1          & 67.4          & 63.5          & 63.0                          & 73.8          \\
PointNet \cite{qi2017pointnet}   & 73.3          & 73.5          & 72.7          & 68.2          & 68.2                          & 79.2          \\
SpiderCNN \cite{xu2018spidercnn} & 77.1          & 78.1          & 77.7          & 73.8          & 73.7                          & 79.5          \\
PointNet++ \cite{qi2017pointnet++} & 82.3          & 82.7          & 81.4          & 79.1          & 77.9                          & 84.3          \\
DGCNN  \cite{wang2019dynamic}    & 82.8          & 83.3          & 81.5          & 80.0          & 78.1                          & 86.2          \\
PointCNN \cite{li2018pointcnn}   & 86.1          & 83.6          & 82.5          & 78.5          & 78.5                          & 85.5          \\ \hline
ODFNet     & \textbf{87.2} & \textbf{88.9} & \textbf{86.7} & \textbf{88.8} & \textbf{85.1}                 & \textbf{89.3}
\end{tabular}
\end{table*}

\begin{table*}[]
\centering
\resizebox{\textwidth}{!}{%
\begin{tabular}{l|l|ll|llllllllllllllll}
\hline
Method & input & mpIoU & mIoU & a.plane & bag & cap & car & chair & e.phone & guitar & knife & lamp & laptop & m.bike & mug & pistol & rocket & s.board & table \\ \hline
PointNet \cite{qi2017pointnet} & p & 80.4 & 83.7 & 83.4 & 78.7 & 82.5 & 74.9 & 89.6 & 73.0 & 91.5 & 85.9 & 80.8 & 95.3 & 65.2 & 93.0 & 81.2 & 57.9 & 72.8 & 80.6 \\
PointNet++ \cite{qi2017pointnet++} & p+n & 81.9 & 85.1 & 82.4 & 79.0 & 87.7 & 77.3 & 90.8 & 71.8 & 91.0 & 85.9 & 83.7 & 95.3 & 71.6 & 94.1 & 81.3 & 58.7 & 76.4 & 82.6 \\
DGCNN \cite{wang2019dynamic} & p & \textbf{84.6} & 85.2 & 84.0 & 83.4 & 86.7 & 77.8 & 90.6 & 74.7 & 91.2 & 87.5 & 82.8 & 95.7 & 66.3 & 94.9 & 81.1 & 63.5 & 74.5 & 82.6 \\
PCNN \cite{atzmon2018point} & p & 81.8 & 85.1 & 82.4 & 80.1 & 85.5 & \textbf{79.5} & 90.8 & 73.2 & 91.3 & 86.0 & 85.0 & 95.7 & 73.2 & 94.8 & \textbf{83.3} & 51.0 & 75.0 & 81.8 \\
DensePoint \cite{liu2019densepoint} & p & 84.2 & 86.4 & 84.0 & \textbf{85.4} & \textbf{90.0} & 79.2 & 91.1 & \textbf{81.6} & 91.5 & 87.5 & 84.7 & 95.9 & 74.3 & 94.6 & 82.9 & \textbf{64.6} & 76.8 & \textbf{83.7} \\
Point2Sequence \cite{liu2019point2sequence} & p & 82.2 & 85.2 & 82.6 & 81.8 & 87.5 & 77.3 & 90.8 & 77.1 & 91.1 & 86.9 & 83.9 & 95.7 & 70.8 & 94.6 & 79.3 & 58.1 & 75.2 & 82.8 \\ \hline
ODFNet & p & 83.3 & \textbf{86.5} & \textbf{85.1} & 85.0 & 89.5 & 78.7 & \textbf{91.9} & 73.6 & \textbf{92.2} & \textbf{88.2} & \textbf{85.9} & \textbf{96.1} & \textbf{74.9} & \textbf{95.3} & 82.2 & 53.7 & \textbf{77.7} & 83.3 \\ \hline
\end{tabular}%
}
\caption{ShapeNet Part segmentation results for different architectures. Input column indicates whether points (p) and normals (n) are used.}
\label{table:shapenet}
\end{table*}

\begin{figure}
\begin{center}
%\fbox{\rule{0pt}{2in} \rule{.9\linewidth}{0pt}}
   \includegraphics[width=\linewidth]{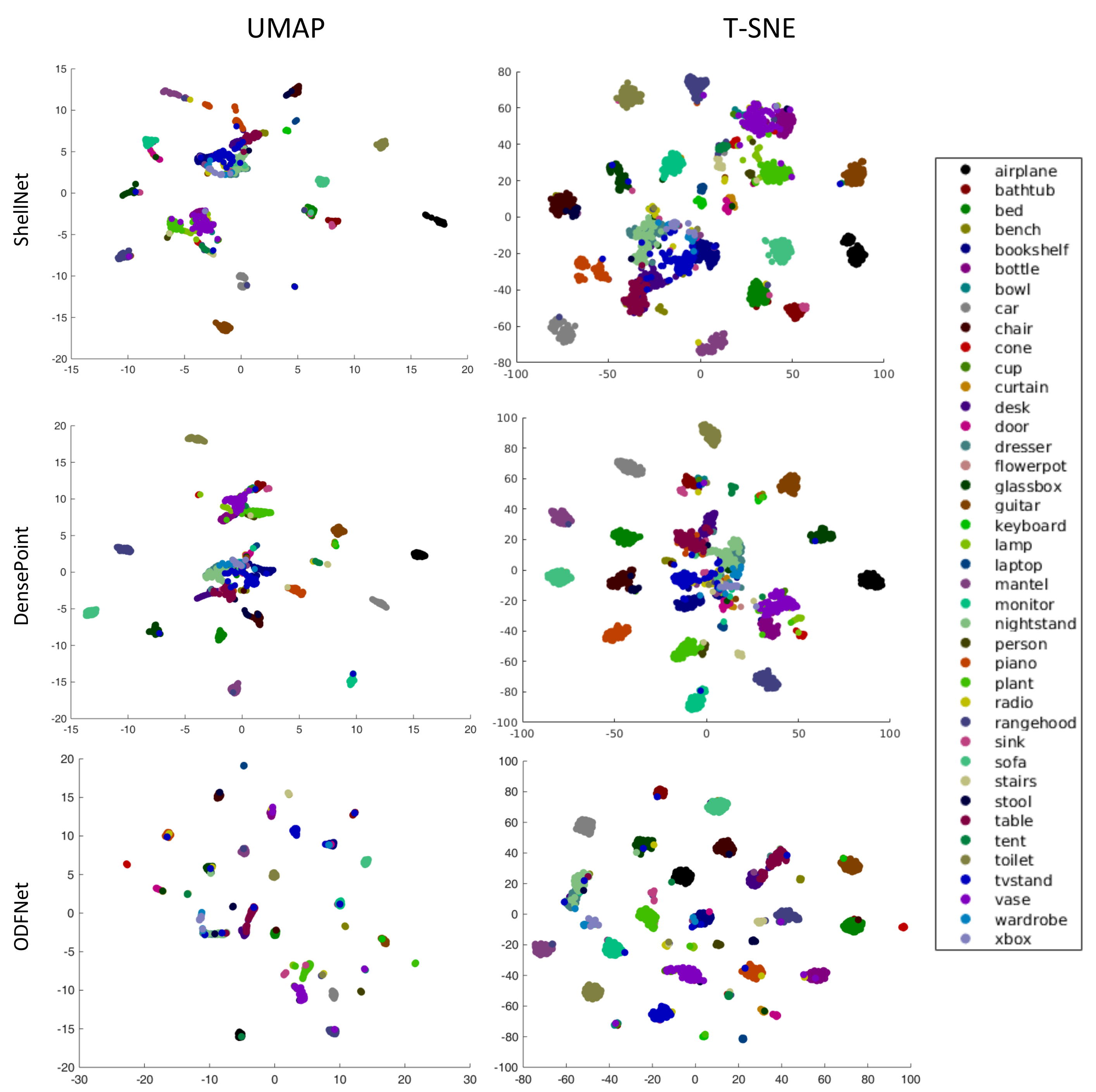}
\end{center}
   \caption{The projections from the last layer outputs of ODFNet, DensePoint and ShellNet architectures for ModelNet40 dataset onto a 2D space using the UMAP \cite{mcinnes2018umap} and t-SNE \cite{maaten2008visualizing} methods.
   }
\label{fig:UMAP}
\end{figure}

\textbf{Further Experiments: } To further investigate our network's capacity for point cloud object abstraction, for each test subject, we extract the output of the last classification layer for the ODFNet, as well as for DensePoint and ShellNet, where the latter two are the previous SoTA point cloud architectures. Then, using two widely utilized dimensionality reduction techniques, UMAP \cite{mcinnes2018umap} and t-SNE \cite{maaten2008visualizing}, we project those vectors onto 2D space and assess this mapping by the Silhouette score \cite{rousseeuw1987silhouettes}, which evaluates whether each object is well matched to its own cluster. The scores are given in Table \ref{table:Silhouette}, and the projections are visualized in Figure \ref{fig:UMAP}. As can be seen from the results, although quantitative scores on the classification of these methods are close to each other, ODFNet's features appear more distinctive and produce a relatively more separated layout than the other two methods, which is also observed in the Silhouette scores.

\begin{table}[]
\centering
\caption{Silhouette scores for last layers of ODFNet, DGCNN and ShellNet after UMAP \cite{mcinnes2018umap} and t-SNE \cite{maaten2008visualizing} projection. \{worst:best\}:\{-1:1\}}
\label{table:Silhouette}
\begin{center}
\begin{tabular}{l|cc}
\hline
Method   & tSNE S. Score & UMAP S. Score\\ \hline
ODFNet   & \textbf{0.623} & \textbf{0.660} \\
DensePoint  & 0.453 & 0.474 \\
ShellNet & 0.472 & 0.466 \\ \hline
\end{tabular}
\end{center}
\end{table}

\begin{figure}[]
\begin{center}
  \includegraphics[width=\linewidth]{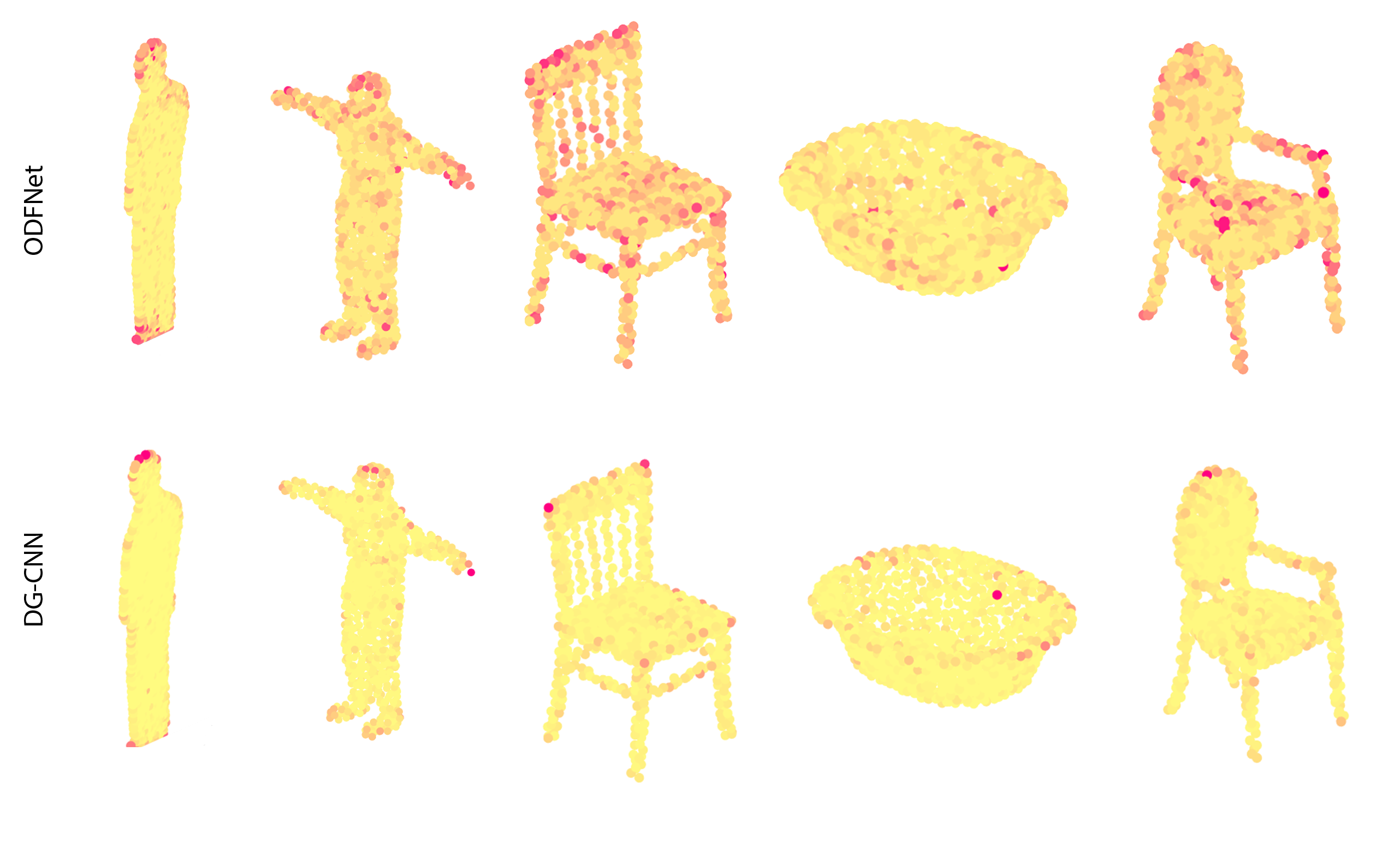}
\end{center}
  \caption{Different heat maps for DGCNN and ODFNet.}
\label{fig:heatmaps}
\end{figure}

Furthermore, to understand which points generate global features at the classification step, features before the last max-pooling of the network are examined. Heat maps according to the contribution of each point to the final result are obtained for ODFNet and DGCNN as given in Figure \ref{fig:heatmaps}. It can be observed that the ODFNet selects more diverse and seemingly important feature points from the point clouds when compared to the DGCNN\footnote{Pointwise heat maps cannot be created for DensePoint and ShellNet since they do not use directly the points but their distribution.}. According to the visual results, we can conclude that for objects with relatively more complex geometric shapes, visually representative points, which are mostly corners and endpoints, are selected by the ODFNet. The ODFs for those points that are around corners and edges have anisotropic distributions while the ODFs on the flat areas have isotropic distributions with similar strengths over many directions.

We also further investigate our method to analyze the effects of two important decisions we make when calculating ODF values: number of different conic neighborhood directions and pivot selection. Table \ref{table:ablation_feature} shows the accuracies for different settings. The results for Experiment A, B, and C indicate that 42 directions which are obtained by the second tessellation of the icosahedron give the best performance compared to the first (12 directions) and third (162 directions) tessellations. The results for the remaining experiments show that choosing pivots that are rotation-invariant in the x-y plane works better compared to the pivots that are fully rotation-invariant. Because all the training data is aligned according to the z-direction, using this direction as a pivot improves the performance.

\begin{table}[]
\caption{An experiment on selection of hyperparameters: direction count and pivot selection. OA refers to overall accuracy. }
\label{table:ablation_feature}
\centering
\begin{tabular}{
l |
l 
c |
l }
\hline
Experiment & Dir. Count & Pivot Sel.  & OA \\ \hline
A     & 12         & RI-XY               &  92.9    \\
B     & 42         & RI-XY               & 93.4 \\
C     & 162        & RI-XY               &   93.1   \\ \hline
D     & 12        & RI-XYZ               &   91.6   \\ 
E     & 42         & RI-XYZ              &  91.8 \\
F     & 162         & RI-XYZ              &  91.4 \\
\hline
\end{tabular}
\end{table}

\subsection{Part Segmentation}
\label{subsection:shapenet}
We evaluated our segmentation network on the ShapeNet part segmentation benchmark \cite{yi2016scalable}, which contains 14007 training and 1874 test samples, 16 object categories each of them partitioned into $\{2-6\}$ parts, making a total of 50 parts. Following the practice of the previous state-of-the-art \cite{liu2019densepoint}, which used ensembling, during the testing phase, for every test object, we also obtain scaled versions of test objects by $+0.3\%$,$-0.3\%$ in each direction and averaged their scores. We compare our performance with those studies using a point cloud structure. Our experimental results for ShapeNet part segmentation are reported for ODFNet along with previous methods in Table \ref{table:shapenet}.

\subsection{Scene Segmentation}

For scene segmentation, the commonly used S3DIS dataset \cite{armeni20163d} is utilized. The dataset contains point clouds sampled from six different challenging scenes. General practice in experimentation with this dataset involves training with a leave-one-out cross-validation (6-fold) strategy. State-of-the-art results are obtained for S3DIS for the overall accuracy measure as shown in Table \ref{table:S3DISAcc}.

\begin{table}[h]
\centering
\caption{Overall accuracies and mIOU values for point cloud scene segmentation results on S3DIS dataset.}
\label{table:S3DISAcc}
\begin{tabular}{lll}
\hline
Method                                                    & OA & mIoU    \\ \hline
PointNet \cite{qi2017pointnet}           & 78.6 & 47.6 \\
PointNet++ \cite{qi2017pointnet++}       & 81.0 & 54.5 \\
PointSIFT \cite{jiang2018pointsift}      & 88.7 & 70.2 \\
Engelmann \cite{engelmann2018know}    & 84.0 & 58.3 \\
3DContextNet \cite{zeng20183dcontextnet} & 84.9 & 55.6 \\
PointWeb \cite{zhao2019pointweb}         & 87.3 & 66.7 \\
ShellNet \cite{zhang2019shellnet}        & 87.1 & 66.8 \\
PointCNN \cite{li2018pointcnn}           & 88.1 & 65.4 \\
InterpCNN \cite{mao2019interpolated}     & 88.7 & 66.7 \\
DGCNN \cite{wang2019dynamic}             & 84.1 & 56.1 \\
Liu et al. \cite{liu2020self}                       & 88.5 & 64.1 \\
RandLA-Net \cite{hu2020randla}                                              & 88.0 & 70.0 \\
HEPIN \cite{jiang2019hierarchical}                                                    & 88.2 & 67.8 \\
PointWeb \cite{zhao2019pointweb}                                                 & 87.3 & 66.7 \\
CF-SIS \cite{wen2020cf}                                                    & 88.0 & \textbf{74.0} \\ \hline
ODFNet                                                    & \textbf{90.8} & 72.2 \\ \hline
\end{tabular}
\end{table}

\subsection{Ablation Studies}

We perform ablation studies on the ModelNet40 dataset for the classification task to further analyze our decisions for ODFNet and ODF-xyz.

To quantify the functionality of different modules of the ODFNet, we experimented with removing ODF-Dir and ODF-Glob blocks. For these experiments, we use 42 conic neighborhood directions and the RI-XY pivot selection method. The performance results in Table \ref{table:ablation_network} show that both ODF-Dir and ODF-Glob blocks contribute to the performance.

\begin{table}[]
\caption{Ablation study results on ODFNet. OA refers to overall accuracy.}
\label{table:ablation_network}
\centering
\begin{tabular}{
l |
l 
l
l }
\hline
Experiment & ODF-Dir & ODF-Glob & OA \\ \hline
A     & \checkmark         &\checkmark  &  93.4    \\
B     & \checkmark         &            & 92.3 \\
C     &                    &\checkmark  & 91.9     \\
D     &                    &            & 91.2     \\
\hline
\end{tabular}
\end{table}

For the ODFNet-xyz, another ablation study is carried out in order to analyze the two factors contributing to the rotation invariance: (i) using rotation invariant feature vectors instead of difference vectors; and (ii) using RI-XYZ, that is selection of pivot directions that are also rotation invariant, instead of RI-XY. The results given in Table \ref{table:ablation_network2} show that by using both factors, the best accuracy is obtained.

\begin{table}[]

\caption{Ablation study results on ODFNet-xyz. OA refers to overall accuracy.}
\label{table:ablation_network2}

\begin{tabular}{l|ll|lll|l}
\hline
Exp. & \begin{tabular}[c]{@{}l@{}}Rot.Inv.\\ Features\end{tabular} & \begin{tabular}[c]{@{}l@{}}RI-\\ XYZ \end{tabular} & z/z  & \multicolumn{1}{c}{SO3/SO3} & z/SO3 & std. \\ \hline
E          & \checkmark                                                               &        & 90.4 & 85.0                        & 42.9  & 26.0  \\
F          &                                                                 & \checkmark      & 91.6 & 89.5    & 24.9  & 37.9  \\
G          & \checkmark                                                               & \checkmark      & 90.2 & 90.2                        & 90.2  & 0    \\ \hline
\end{tabular}
\end{table}

\section{Discussions and Conclusion}
Our experimental results demonstrate that ODFNet achieves the SoTA performance in both the classification task and particularly for the more challenging segmentation task in ShapeNet and S3DIS. This provides evidence to our hypothesis that the ODF representation, which exploits the idea of incorporation of local point orientation distribution characteristics into the point cloud neural network models, is highly beneficial. 
    
The point orientation distribution features help the neural network models capture further informative characteristics of the point cloud. This is revealed by the attention that the ODFNet model pays to the unique identifying points on an object such as corners, tips, and borders between planar regions. Features generated by the ODFNet correlate with an increased representation power for point clouds.

Moreover, we investigate the rotation invariance properties of ODF representations through a variant of the proposed ODFNet: ODFNet-xyz, which is fully rotation-invariant. Our experimental results show that ODF features can be effectively calculated in a rotation-invariant manner, and ODFNet-xyz performs comparably against state-of-the-art rotation-invariant point cloud analysis models.

From the results in Table \ref{table:MODELNETScoreTable} and \ref{table:rotations}, we can conclude that the original point locations and other location-based properties are highly discriminative. However, there is a trade-off between exploiting rotation-invariant features versus location-based rotation-variant point features, as location-based features are  naturally not robust to rotation changes. Indeed, when the objects are rotated, the performances of rotation-variant networks drastically decrease. Moreover, the established benchmarks and procedures for evaluations on those benchmarks do not reward the rotation-invariant representations, as indicated by the inferior performance of rotation-invariant methods on those benchmarks.

As for future work, in geometric representation learning of point clouds, capturing essential defining characteristics of an object, whether through better augmented definitions of local patches around points as the ODFNet does or in other similarly effective ways of local and global encoding schemes, could provide further improvements in supervised and unsupervised learning tasks on point clouds.

\bibliographystyle{cag-num-names}
\bibliography{refs}

\end{document}